\pdfoutput=1
\documentclass[11pt]{article}

\usepackage[final]{acl}

\usepackage{times}
\usepackage{latexsym}

\usepackage[T1]{fontenc}
\usepackage[utf8]{inputenc}

\usepackage{microtype}

\usepackage{inconsolata}

\usepackage{graphicx}

\usepackage{synttree}
\usepackage{subcaption}
\usepackage{tikz-dependency}
\usepackage[cjk]{kotex}
\usepackage{langsci-gb4e}

\usepackage{graphicx}
\usepackage{inconsolata}
\usepackage{amsmath} 
\usepackage{tabularx}
\usepackage{arydshln}
\usepackage{graphicx}
\usepackage{tikz-dependency}
\usepackage{colortbl}

\usepackage{subcaption}

\newcommand*{\yc}[1]{{\color{black}#1}}
\newcommand*{\ycq}[1]{{\color{black}#1}}

\title{Unlocking Korean Verbs: \\
A User-Friendly Exploration into the Verb Lexicon}

\author{
Seohyun Song$^{1*}$~~~  %alexalex225225@seoultech.ac.kr
Eunkyul Leah Jo$^{2*}$~~~
Yige Chen$^{3*}$~~~
Jeen-Pyo Hong$^{4}$\thanks{Equally contributed authors. $\dagger$Corresponding authors: KyungTae Lim and Jungyeul Park.}~~
Kyuwon Kim$^{5}$\\
{\bf Jin Wee$^{6}$}~~~ %wj0119@korea.kr
{\bf Miyoung Kang$^{6}$}~~~ %mykang@korea.kr
{\bf KyungTae Lim$^{7\dagger}$}~~~
{\bf Jungyeul Park$^{2\dagger}$}~~~
{\bf Chulwoo Park$^{8}$}\\
$^{1}$Seoultech, South Korea~~
$^{2}$The University of British Columbia, Canada \\
$^{3}$The Chinese University of Hong Kong, Hong Kong~~
$^{4}$42dot Inc., South Korea\\
$^{5}$Seoul National University, South Korea~~
$^{6}$National Institute of Korean Language, South Korea\\
$^{7}$KAIST, South Korea~~
$^{8}$Anyang University, South Korea\\
\url{https://www.korean.go.kr}
}

\begin{document}
\maketitle

\begin{abstract}
The Sejong dictionary dataset offers a valuable resource, providing extensive coverage of morphology, syntax, and semantic representation. This dataset can be utilized to explore linguistic information in greater depth.
The labeled linguistic structures within this dataset form the basis for uncovering relationships between words and phrases and their associations with target verbs. 
This paper introduces a user-friendly web interface designed for the collection and consolidation of verb-related information, with a particular focus on subcategorization frames. Additionally, it outlines our efforts in mapping this information by aligning subcategorization frames with corresponding illustrative sentence examples.
Furthermore, we provide a Python library that would simplify syntactic parsing and semantic role labeling. These tools are intended to assist individuals interested in harnessing the Sejong dictionary dataset to develop applications for Korean language processing. 
\end{abstract}

\section{Introduction}

When examining the verb lexicon for Korean, the Sejong dictionary has been a major language resource. 
% Previous research on 
As part of the 21st Century Sejong Project, which aims to advance Korean language information processing, the Sejong dictionary has produced extensive datasets that describe Korean lexicon data in great detail. Although it is designed with a vast amount of morphological, semantic, and syntactic information, there have been minimal attempts to utilize its resources. This paper investigates the Korean verb lexicon, with a particular focus on the subcategorization frames in the Sejong verb dictionary. The objective is to facilitate easier access to and manipulation of the Sejong dictionary, thus promoting further studies and applications in Korean language processing.
% the subcategories a verb requires 

In annotating corpora with semantic roles, the methodology of Proposition Bank (PropBank) has been widely adopted in natural language processing \citep{palmer-gildea-kingsbury:2005:CL}. Similar efforts can be observed in the Sejong dataset, which seeks to provide a meaningful level of linguistic representation and a corpus of annotated data to facilitate further research and analysis. This structured dataset will serve as a foundation for identifying relationships between words, such as organizing and searching for verbs that share the same subcategorization frames.

\begin{figure*}[!ht]
  \centering
\resizebox{\textwidth}{!}
{
  \includegraphics{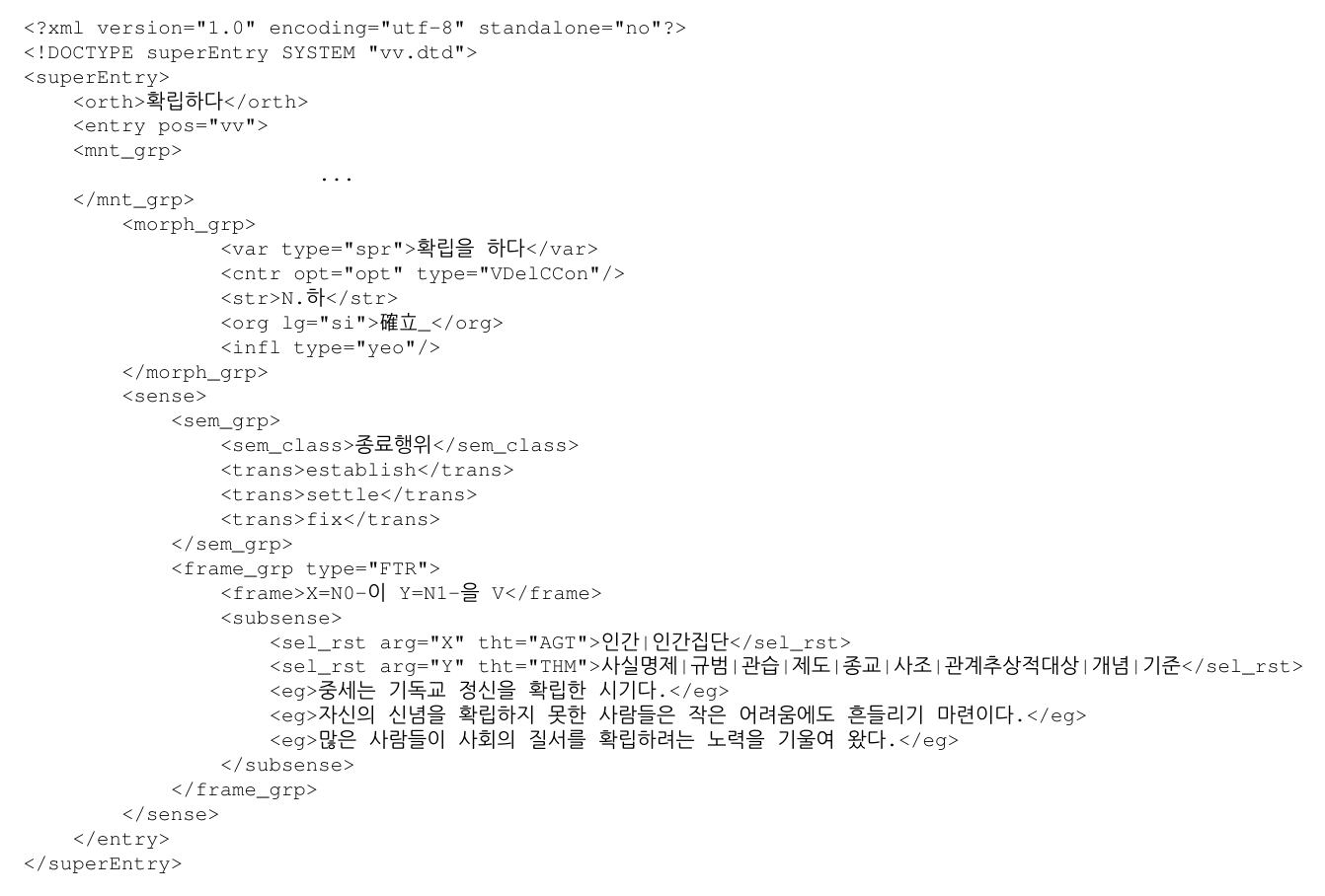}
}
  \caption{Example from the Sejong dictionary: 확립하다\textit{hwaglibhada} (`establish')}
  \label{sjdic-example}
\end{figure*}

Within the Sejong dictionary are XML files for each verb {(Figure~\ref{sjdic-example})}. Inside are XML tags that identify each word's morphological, semantic, and syntactic information with additional metadata. 
Some annotations make it possible to identify {arguments such as} semantic roles of verbs in sentences {(Figure~\ref{sjdic-prjection})}. 
A set of the underlying semantic roles for each verb and frame groupings are defined for each sense of the verb. Then, in each group are the frames that annotate selection restrictions with example sentences. We believe this detailed verb representation level is vital for many language processing applications. 
% With such hand-annotated information 
With thorough and manual documentation on subcategorization in the Sejong dictionary, we aim to enhance the dataset's usability, facilitating the creation of user-friendly systems for all future projects.

\begin{figure*}[!ht]
  \centering
\footnotesize  % \includegraphics{}
\begin{tabularx}{\textwidth}{r XXX}
Frame: & \texttt{\colorbox{blue!30}{X=N0-이}} &
\texttt{\colorbox{red!30}{Y=N1-을}} & \texttt{V} \\
& {\texttt{arg="\colorbox{blue!30}{X}" tht="AGT"}} & 
{\texttt{arg="\colorbox{red!30}{Y}" tht="THM"}} & \\
Sentence: & [\colorbox{blue!30}{많은 사람들이}]$_{\texttt{AGT}}$ &
[\colorbox{red!30}{사회의 질서를}]$_{\texttt{THM}}$ & 
[확립하려는]$_{\texttt{TARGET}}$ ~~~... \\
& \textit{manh-eun salamdeul-i} & \textit{sahoeui jilseoleul} & 
\textit{hwaglibhalyeoneun ~~~...} \\
& {Many people.\texttt{nom}} & {social order.\texttt{acc}} & {establish} \\
& \multicolumn{3}{c}{`Many people have made efforts to establish social order.'}
\end{tabularx}
% [Figure: Frame to sentence projection example]  
  \caption{Example of subcategorziation frame to sentence projection}
\label{sjdic-prjection}
\end{figure*}

In this paper, we introduce a web interface created for individuals with minimal programming experience or those seeking quick access to the verbs in the Sejong dictionary. This interface consolidates verb information and provides easy access to subcategorization frames. It also includes our efforts to map information by projecting subcategorization frames onto corresponding sentence examples.
Additionally, we provide a Python library that includes tools for syntactic parsing and semantic role labeling, designed for those who wish to utilize the Sejong dictionary in developing Korean language processing applications.
Therefore, our main contribution is two-fold: First, we offer a corpus of sentence examples from the Sejong dictionary with annotated argument-predicate boundaries. Second, we develop two key tools—a web interface and programming libraries—to facilitate access to Sejong dictionary information.

\section{Previous work}

\subsection{Subcategorization frames}
Subcategorization refers to the tendency that a lexeme, usually a verb, has its restrictions regarding the syntactic arguments it takes. To address such tendencies, subcategorization frames are provided as the formalization of subcategorization, explicitly describing the arguments the verbs may select. 
There are two major {studies} of subcategorization frames for Korean in addition to the Sejong dictionary: \yc{(1)} manually defined subcategorization frames in the context of tree-adjoining grammars \citep{han-EtAl:2000}, and \yc{(2)}
automatically extracted subcategorization frames from the treebank \citep{xia-EtAl:2000:CLPW,park:2006}. Subcategorization frames have been linguistically studied thoroughly \citep{park-2002-criteria,younghee-2004-criterion}, mostly for distinguishing between arguments and adjuncts in Korean within subcategorization frames.

Before presenting our method for improving access to verb subcategorization, we will briefly review previous projects that describe Korean grammar, each specifying a distinct subcategorization frame. Additionally, two key Korean language resources provide verb-related information: the Korean Proposition Bank and the National Institute of Korean Language (NIKL) Semantic Role Labeling (SRL) dataset. To assess and compare the scope of these resources, we conducted a quantitative analysis by examining the number of verbs and their associated frames, as detailed in Table~\ref{comparison}.
Korean Proposition Bank (Korean PropBank)\footnote{\url{https://catalog.ldc.upenn.edu/LDC2006T03}} has achieved the same level of recognition and development as its English counterpart, Penn's Proposition Bank \citep{palmer-gildea-kingsbury:2005:CL}. 
As resources, Proposition Banks provide information about the predicate-argument structure of sentences in a corpus.
Similarly, Korean PropBank provides annotated sentences from the Penn Korean treebank \citep{han-EtAl:2002}.
The NIKL SRL also provides detailed verb information in addition to the SLR corpus. 
It refines 1,597 verbs from the previous Sejong dictionary's entry and adds 2,063 new verbs along with their subcategorization frame information. Interestingly, it follows the annotation scheme of Penn's Propbank with the Sejong treebank labels. 
As the name suggests, its primary goal is to provide the SRL annotated corpus for Korean.

\begin{table}[!ht]
\centering
\footnotesize{
\begin{tabular}{r |cc c} \hline 
% \hlineB{4}
&  {Sejong}  & {K-PropBank} & {NIKL}\\ \hline
\# of verbs  & 15,181  & 2749  & 3660\\
av. \# of frames & 1.812 & 1.408 & 1.593\\ \hline
% \hlineB{4}
% \# of arguments && \\
\end{tabular}
}
  \caption{Basic statistics: number of verbs and their average number of frames per verb}
  \label{comparison}
\end{table}

\subsection{Previous usages of the Sejong dictionary}
%(Yige) korean srl\footnote{\url{https://korean.go.kr}} / framenet\footnote{\url{https://github.com/machinereading/koreanframenet}}

The Sejong dictionary has been adopted in a limited way. Data and example sentences from the Sejong dictionary were used in the Korean SRL dataset developed by the National Institute of Korean Language, which was later improved through a linguistically-informed annotation strategy \citep{chen-etal-2024-linguistically-informed}. Additionally, a small selection of example sentences from the Sejong dictionary was incorporated into Korean FrameNet \citep{hahm-EtAl:2018:LREC,chen-etal-2024-towards-standardized}, guided by the frames outlined in the dictionary entries  within the Sejong dictionary.\footnote{\url{https://github.com/machinereading/koreanframenet}}$^{,}$\footnote{\url{https://github.com/jungyeul/k-framenet}}

\section{Developing a Web  Interface}

We offer a user-friendly interface that allows for easy navigation and access to the Sejong dictionary, catering to individuals with non-computational backgrounds. Visually, words are organized in Korean alphabetical order, enabling users to select each word to examine its linguistic details. A particular piece of critical information that we decided to include is example sentences illustrating the identified argument boundary for each word. To complement this feature, we can expand the service for searching words by their frame, argument, morphological inflection and semantic group. This new systematic approach and visual aid in organizing the Sejong dictionary can help many users understand the relationships between words more intuitively. % yige: I put this sent here directly after this paragraph which I found a bit smoother

Data in the Sejong dictionary is organized so that for each verb lexeme, a set of its morphological, syntactic, and semantic information is provided along with example sentences. The morphological information is usually represented as inflections given the agglutinative feature of Korean, and the dataset specifies whether the inflections are regular. Furthermore, the Sejong dictionary distinguishes different senses for each lexeme. Under each of its possible senses, the dictionary offers both the verb category and a translation of the sense, along with the corresponding frame that indicates the arguments required by the verb.

\begin{figure}[!ht]
\centering
\resizebox{.45\textwidth}{!}
{
\includegraphics{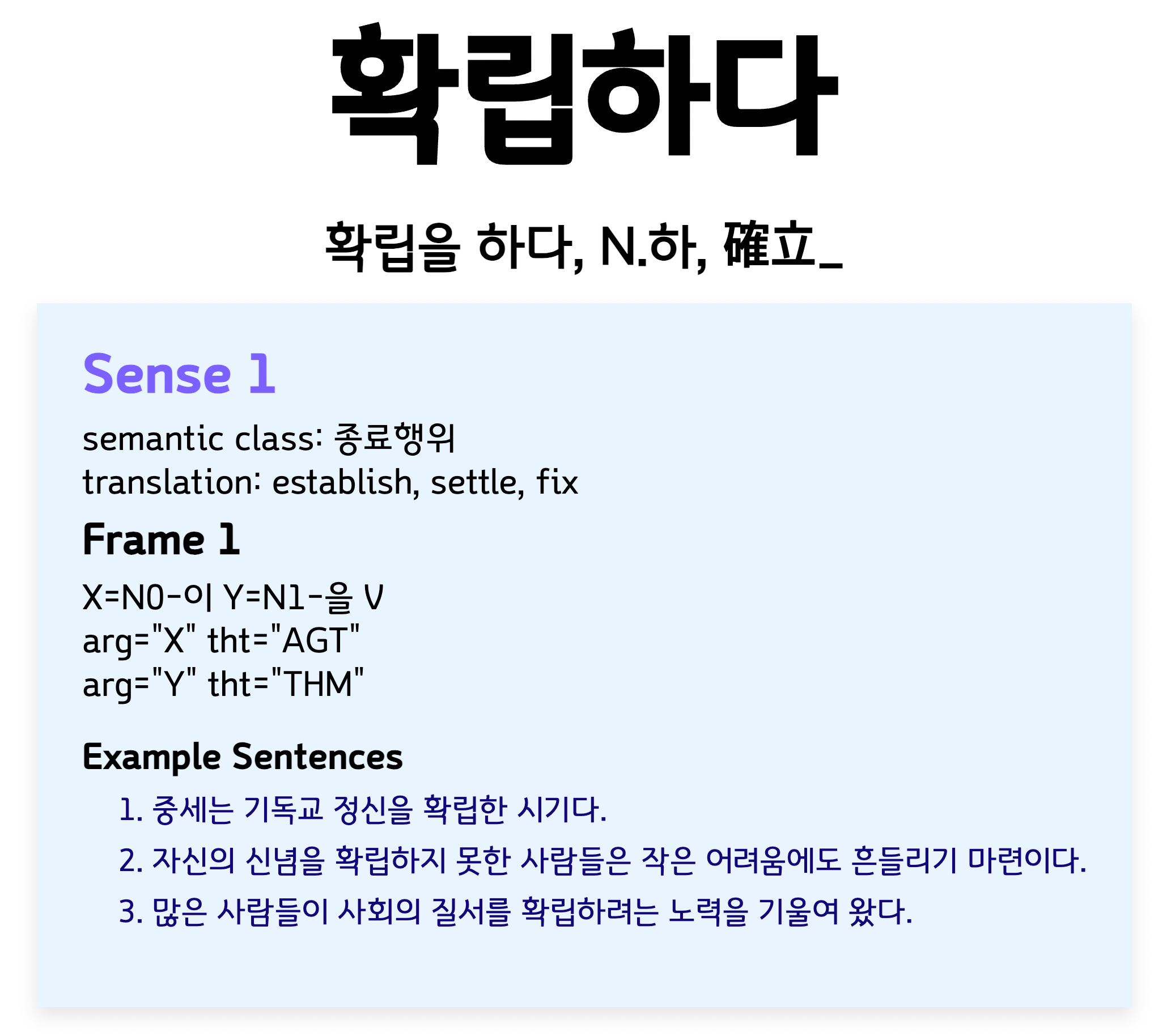}
}
\caption{Interface screenshot of a typical word details page including morphological, semantic, and syntactic frame information along with their associated semantic roles and sentence examples.}
  \label{sceenshot1}
\end{figure}

Figure \ref{sceenshot1} is a screenshot of our interface upon displaying the word 확립하다 \textit{hwaglibhada} (`establish'). We obtain the word's information from the Sejong dictionary's XML file. The tags that organize the lexicon information are taken from the files when rendering this screen. For example, \yc{the orthographic information of the verb, as shown below, }

{\footnotesize
\begin{verbatim}
  <orth>확립하다</orth>
\end{verbatim}
}
\noindent corresponds with the title of the page. Other \yc{morphological and lexical} tags such as these are also displayed near the top as the subtitle of the word's detail page\yc{, suggesting that the verb can be considered as the combination of a nominal morpheme of Chinese origin and the \ycq{verb-generating derivational suffix 하 (\textit{ha})}}:

{\footnotesize
\begin{verbatim}
  <morph_grp>
  <var type="spr">확립을 하다</var>
  <str>N.하</str>
  <org lg="si">確立_</org>
  </morph_grp>
\end{verbatim}
}

Under the tag sense is the semantic information\yc{, including the semantic class and its English translation,} denoted by \texttt{<sem\_class>} and \texttt{<trans>} \yc{correspondingly}. 
The \yc{subcategorization} frame \yc{of the instance} corresponds with tags under \texttt{<frame\_grp>} and \texttt{<frame>}. Further below are denoted by \texttt{<subsense>} and \texttt{<sel\_rst>} to display the information \yc{with regard to the arguments of the verb}. The semantic roles of the required arguments are introduced in \texttt{<sel\_rst>}, including the general categories such as \textsc{agent} (\texttt{AGT}) and \textsc{theme} (\texttt{THM}), as well as specific categories such as \texttt{인간} \textit{ingan} (`human') and \texttt{사실명제} \textit{sasilmyeongje} (`factual proposition'). The example of the sentences with argument boundaries, marked by \texttt{<eg>} in XML, are displayed as the last section on the details page.
%\ycq{description of semantic info in sjdict}

\begin{figure}[!ht]
  \centering
\resizebox{.45\textwidth}{!}{  
  \includegraphics{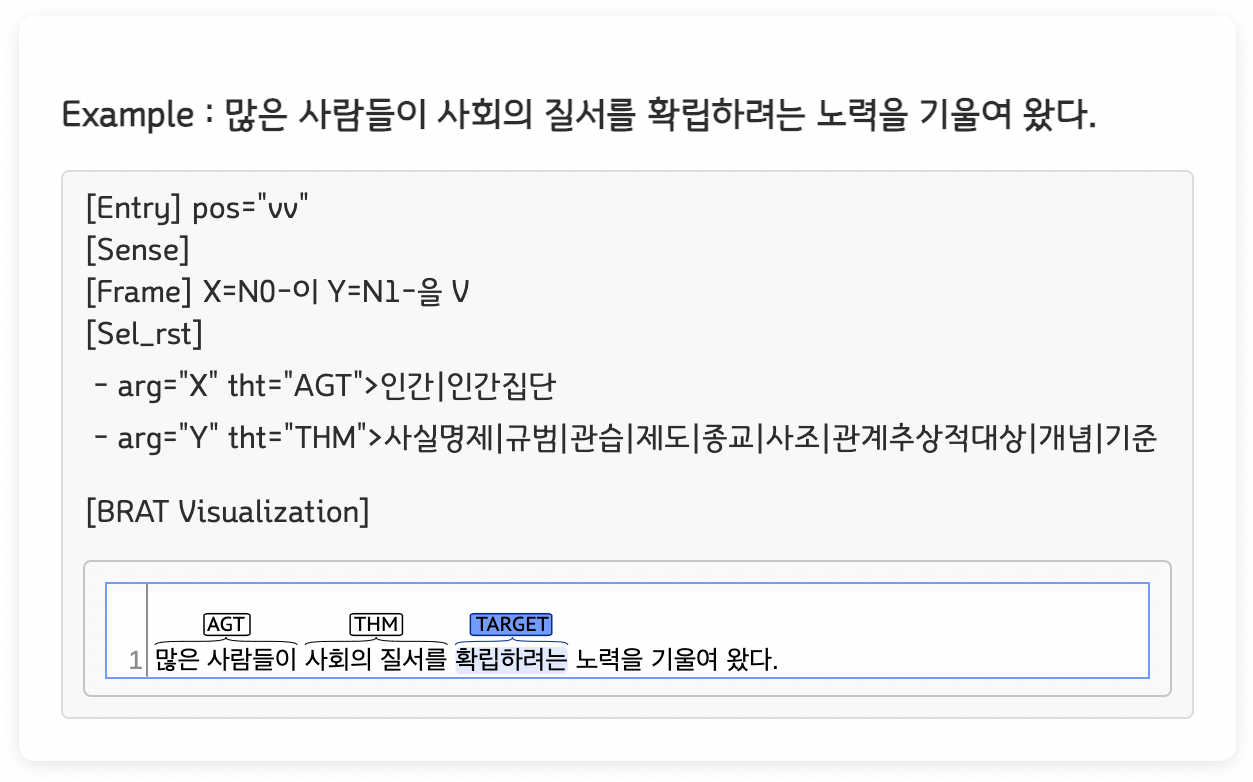}
}
\caption{Subcategorization frames of the verb, along with its semantic role information for arguments, including chunked structures based on frame information to identify argument boundaries}\label{sceenshot3}
\end{figure}

In the web interface, we also introduce the distinction of sentence argument boundaries for the sentence examples. These argument boundaries are determined based on the frame information of the provided verb in the Sejong dictionary, as shown in Figure~\ref{sceenshot3}. The argument boundaries are presented using \texttt{brat} \citep{stenetorp-etal-2011-bionlp}\footnote{\url{https://brat.nlplab.org}} for the following sentence:

\begin{center}
\resizebox{.5\textwidth}{!}{
{\footnotesize
\begin{tabular}{l l l l}
$[$많은 사람들이$]$ &  $[$사회의 질서를$]$ & $[$확립하려는$]$ & ...     \\
$_{\textsc{agent}}$ & $_{\textsc{theme}}$ & $_{\textsc{target}}$ &  \\
\textit{manheun salam-deul-i} & \textit{sahoe-ui jilseo-leul} & \textit{hwaglibha...} &  \textit{...}\\
many person-\textsc{pl}-\textsc{nom} & society-\textsc{gec} order-\textsc{acc} & establish & ... \\
\multicolumn{4}{l}{(`Many people have made efforts to establish order in society.')}\\
% {\tiny ~~}~&&&\\
\end{tabular}
}}
\end{center}
\noindent  where the frame suggests that the predicate requires two arguments: an agent and a theme. Consequently, these roles are assigned to the detected arguments in the example sentence.

% \begin{exe}
% \ex \gll 많은 사람-들-이 사회-의 질서-를 확립-하려-는 노력-을 기울이-어 왔-다 \\
%      many person-PL-NOM society-GEN order-ACC establish-try-REL effort-ACC make-CONV come-PST \\
% \trans `Many people have made efforts to establish order in society'
% \end{exe}

\yc{The argument boundaries are detected through a combination of a heuristic-based chunking method and the sentence's dependency structures. Argument boundaries are determined by the locations of postpositions in the sentence, as postpositions in Korean typically indicate the endpoints of arguments and modifiers, based on our previous work \citep{chen-etal-2024-linguistically-informed}. On the other hand, dependency structures assist in identifying argument boundaries when the heuristics are insufficient to derive chunks for certain constituents \citep{chen-etal-2022-yet}. The postpositions of the chunks are further compared with the postpositions given by the frame to determine whether a chunk is an argument or not.}

%%%%%%

\begin{figure}[!ht]
\centering
\resizebox{.45\textwidth}{!}{
  \includegraphics{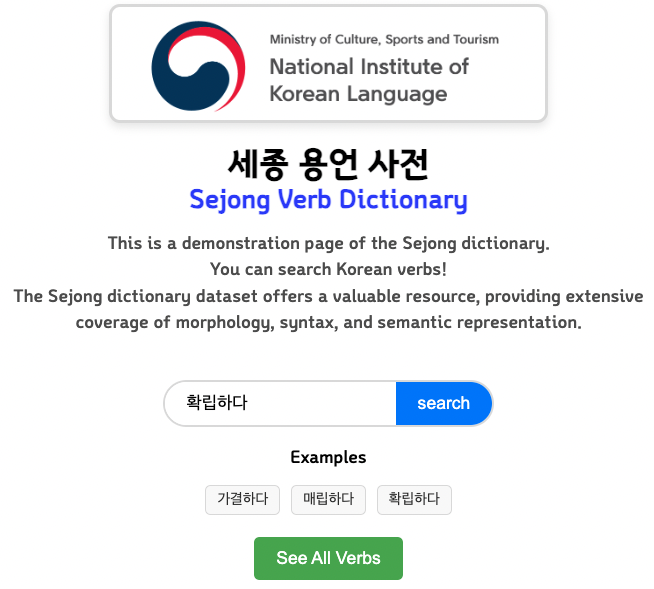}
}
\caption{First page of the web interface, which allows users to search for Korean verbs} \label{sceenshot2}
\end{figure}

The web interface also provides additional functionalities to enable easy navigation for users and linguists, particularly those without a background or expertise in computer science or computing. 
Accordingly, users can swiftly navigate through the dictionary by searching for specific verbs, frames, arguments, morphological inflection and semantic role groups. 
After a search, corresponding words will be listed and clicking upon the button for Sejong will guide the user to a word details page, as in Figure~\ref{sceenshot2}.

\section{Python Library}

This section introduces the \texttt{pySejongFrame}, a Python library designed for processing and accessing the 21st-century Sejong Frame Dictionary. With this tool, users can streamline the manipulation of linguistic data for research and development purposes. This library provides various methods for loading and querying the corpus, with support for both direct and lazy loading techniques. Below, we provide a detailed walk-through of the system’s initialization and querying capabilities, showcasing its practical use for both corpus-based applications and linguistic research.

\subsection{Library initialization}

To begin using the \texttt{pySejongFrame} library, users must set up the environment by specifying paths for the library and data files. The system assumes that the XML data for the Sejong Frame Dictionary is located in the \texttt{share/sejong\_framefiles} directory.

{\footnotesize
\begin{verbatim}
import sys
import os
from pathlib import PurePath, Path
PATH_THIS = PurePath(os.getcwd()).parent
PATH_PYLIB = Path(PATH_THIS, "python")
PATH_SHARE = Path(PATH_THIS, "share")
sys.path.append(str(PATH_PYLIB.resolve()))
\end{verbatim}
}

\noindent
Once the environment is configured, users may create an instance of the \texttt{SejongFrameCorpusReader}.
% , as in $\S$\ref{sec:loadingmethods}.

\subsection{Loading methods}\label{sec:loadingmethods}

The library offers four different methods for loading the Sejong Frame Dictionary. This allows flexibility depending on the user’s requirements. Below, we describe each loading method:

\paragraph{Direct loading using \texttt{SejongFrameCorpusReader}}

In this approach, the corpus is loaded directly using the \texttt{SejongFrameCorpusReader} class.

{\footnotesize
\begin{verbatim}
from SejongFrame.reader \
  import SejongFrameCorpusReader
sejong_framenet = SejongFrameCorpusReader(
  Path(PATH_SHARE, "sejong_framefiles"), "*.xml")
\end{verbatim}
}

\paragraph{Lazy loading with \texttt{LazyClassLoader}}

Lazy loading defers the creation of the index until the corpus is actually accessed. This can save time when dealing with large datasets by delaying the initialization process.

{\footnotesize
\begin{verbatim}
from SejongFrame.loader \
  import LazyClassLoader
sejong_framenet = LazyClassLoader(
  "sejong_framenet", SejongFrameCorpusReader,
  Path(PATH_SHARE, "sejong_framefiles"), "*.xml")
\end{verbatim}
}

\paragraph{Integration with \texttt{nltk}}
The \texttt{pySejongFrame} library can be integrated with \texttt{nltk}, enabling users familiar with the \texttt{nltk} framework toto leverage its tools in conjunction with the Sejong Frame Dictionary.

{\footnotesize
\begin{verbatim}
from SejongFrame.nltk.reader \
  import SejongFrameCorpusReader

sejong_framenet = SejongFrameCorpusReader(
  "sejong_framefiles", 
  r".*\.xml", 
  nltk_data_subdir=str(PATH_SHARE)
)
\end{verbatim}
}

\subsection{Corpus querying and usage}

After loading the Sejong Frame Dictionary, the library allows users to query the corpus for specific linguistic data. The example below demonstrates how to retrieve entries related to the verb {가감하다} (`adjust').

{\footnotesize
\begin{verbatim}
gagam = sejong_framenet.entries("가감하다")
print(repr(gagam))
\end{verbatim}
}
\noindent
Users can also query specific verb forms, such as {수정하다} (`modify') with the tag \texttt{vv}, which refers to the verb form:

{\footnotesize
\begin{verbatim}
sujeong_vv \
  = sejong_framenet.entries("수정하다.vv")

print(repr(sujeong_vv))

# Accessing a specific entry
print(sujeong_vv[2])  
\end{verbatim}
}

\subsection{Exploring Frame Semantics}

The Sejong Frame Dictionary includes rich frame semantics, which allow users to explore semantic classes and frame structures freely. For example, we can examine the semantic group (\texttt{sem\_grp}) and translation of the word {수정하다} by accessing its specific sense:

{\footnotesize
\begin{verbatim}
sense = sujeong_vv['vv.3']['1']

# Semantic group
print(sense.category)  

# Translation
print(sense.trans)  

# Frames within the sense
print(len(sense.frames), sense.frames)  
\end{verbatim}
}
\noindent
Frames also contain arguments, which can be accessed in the following manner:

{\footnotesize
\begin{verbatim}
frame = sense.frames[0]

# Accessing a specific argument
argY = frame.Y  

# Printing the argument in text form
print(str(argY))  
\end{verbatim}
}

\subsection{Error handling and best practices}

While using the \texttt{pySejongFrame} library, users are advised to handle potential errors that may arise due to inconsistent corpus formats or malformed queries. For instance, ensuring that entries exist for the queried lemma or morpheme tag combination can prevent runtime exceptions.

{\footnotesize
\begin{verbatim}
try:
  target = sujeong_vv['vv.3']
except KeyError:
  print("Entry not found for the specified key.")
\end{verbatim}
}
\subsection{Performance considerations}

For large-scale linguistic research, the library’s lazy loading feature ($\S$\ref{sec:loadingmethods}) is highly recommended. In this way, this method optimizes the system's performance by postponing indexing until the data is actively queried. Furthermore, the integration with \texttt{nltk} allows for compatibility with existing NLP workflows, making the system versatile for different use cases.

\subsection{Summary}
The \texttt{pySejongFrame} library provides a toolset for accessing and analyzing the Sejong Frame Dictionary. With multiple loading options, efficient querying mechanisms, and integration with widely-used NLP libraries, it enables researchers to conduct comprehensive analyses of frame semantics in Korean. The library’s ability to work with large datasets and its support for lazy loading make it an ideal choice for corpus-based linguistic studies.

\section{Conclusion and Future Perspectives}

Our current contributions, particularly the Python library, have the potential to be applied across various linguistic tasks, including syntactic parsing systems that utilize lexical information, such as lexicalized parsers \citep{sarkar-han:2002:TAG,park-EtAl:2013:IWPT}, semantic role labeling \citep{changki-etal-2015,seoin:2019,chen-etal-2024-linguistically-informed,chen-etal-2024-misidentified}, and FrameNet parsing \citep{park-EtAl:2014:ISWCPOSTER,kim-etal-2016-korean,hahm-EtAl:2018:LREC,chen-etal-2024-towards-standardized} for the Korean language. We are committed to broadening access to Korean language data beyond traditional boundaries, offering a unified platform for linguists and computer scientists to explore the rich resources of the Korean language. This platform will be invaluable for a wide range of Korean language processing tasks and future linguistic research.

We are continuously working on integrating additional Korean verb lexicons, such as the Korean PropBank and FrameNet, into our web service. Our ultimate goal is to develop a comprehensive Korean VerbNet. By systematically comparing verbs and their subcategorization frames across the Sejong verb dictionary, PropBank, and other resources—and resolving any discrepancies—we aim to harmonize frame information across our interfaces. This will offer users a more cohesive and seamless experience when exploring Korean language data, contributing to advancements in Korean linguistic research and natural language processing.
{Our system, including its Python library with sample verb files, can be accessed at \url{http://117.17.185.27:8025/}.}
\begin{center}
\includegraphics[width=.15\textwidth]{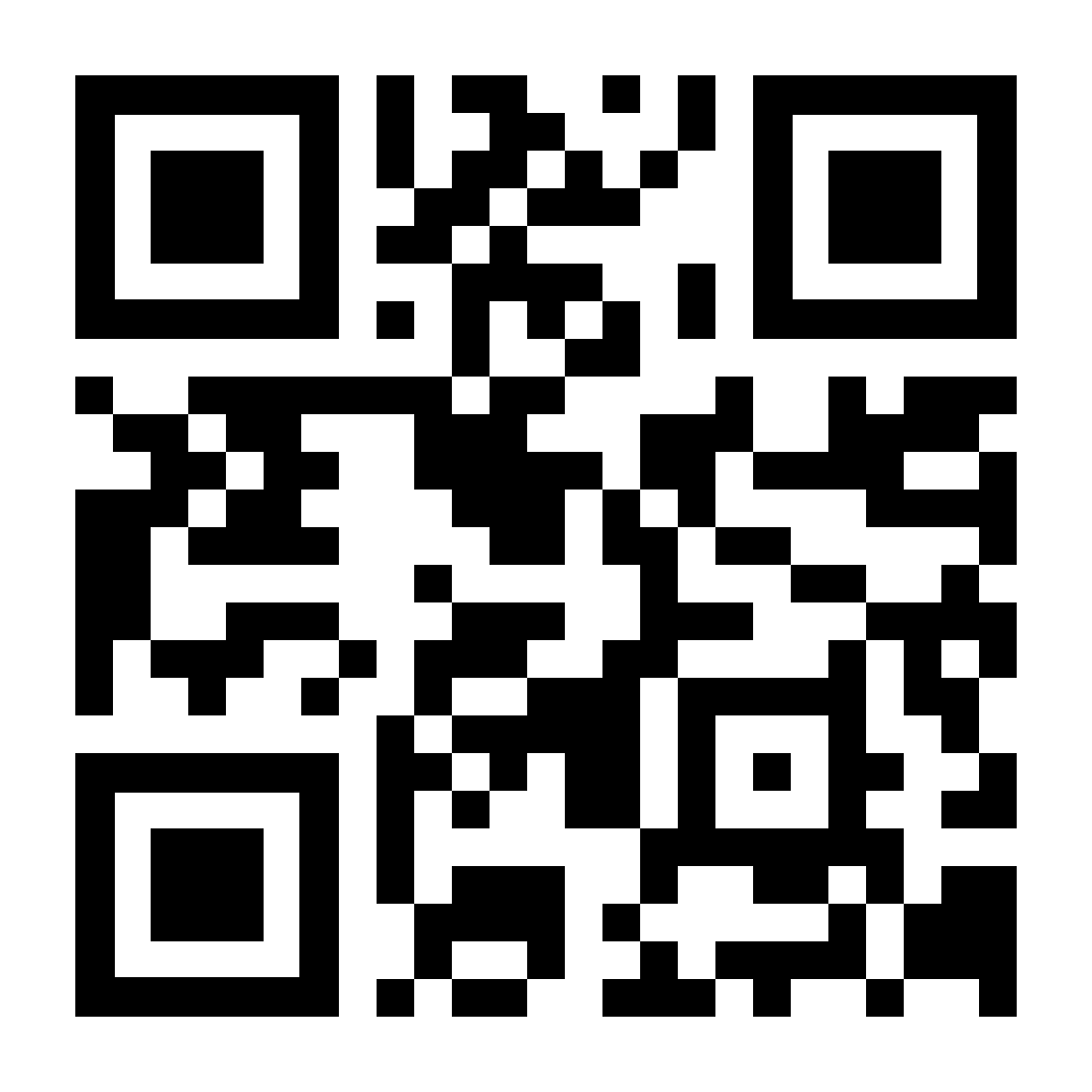}
\end{center}

\section*{Disclaim} 
The Sejong Dictionary of the Sejong Project, was initiated by the Ministry of Culture and Tourism in Korea as part of the 21st Century Sejong Project in 1998. The goal of this initiative was to advance Korean language information processing, resulting in the creation of the largest corpus and language resources for the Korean language.
Previously, third parties were prohibited from distributing content from the Sejong Project, including derivative works like the Python library. However, after successful negotiations with the National Institute of Korean Language, which holds the project’s copyright, we have secured permission to distribute these materials. This milestone has significantly improved accessibility, allowing for broader use of these valuable linguistic resources.
Although the dictionary files required for the Python library must still be obtained directly from the National Institute of Korean Language, the institute will host the web system, ensuring it is available for public use. This arrangement marks an important step forward in making the Sejong Project’s resources more widely accessible to researchers and developers alike.

\section*{Limitation}
One limitation of this work is that, since the dataset is static, the system may not account for the evolution of the language or new linguistic phenomena without regular updates to both the dataset and system functionality.

\section*{Acknowledgments}
This work was supported by Institute of Information \& communications Technology Planning \& Evaluation (IITP) grant, funded by the Korea government (MSIT) (No.RS-2024-00456709, A Development of Self-Evolving Deepfake Detection Technology to Prevent the Socially Malicious Use of Generative AI) awarded to KyungTae Lim.

% \appendix

% \section{Example Appendix}
% \label{sec:appendix}

% This is an appendix.

\end{document}